\documentclass[runningheads]{llncs}
\usepackage[T1]{fontenc}
\usepackage{graphicx}
\usepackage{amsmath}
\usepackage{amssymb}
\usepackage{booktabs}

\usepackage{makecell}
\usepackage{graphicx}
\usepackage{color}
\usepackage{nicefrac}
\usepackage{amsmath}
\usepackage[square,numbers]{natbib}  
\newif\ifdraft
\draftfalse
\drafttrue 

\definecolor{orange}{rgb}{1,0.5,0}
\definecolor{violet}{RGB}{70,0,170}
\definecolor{magenta}{RGB}{170,0,170}
\definecolor{dgreen}{RGB}{0,150,0}

\usepackage[normalem]{ulem}

\ifdraft
\newcommand{\PF}[1]{{\color{red}{\bf PF: #1}}}

\newcommand{\PUW}[1]{{\color{blue}{\bf PUW: #1}}}

\newcommand{\MS}[1]{{\color{dgreen}{\bf MS: #1}}}

\newcommand{\PMJ}[1]{{\color{magenta}{\bf PMJ: #1}}}
\newcommand{\pmj}[1]{{\color{magenta} #1}}
\else
\newcommand{\PF}[1]{}

\newcommand{\PUW}[1]{}

\newcommand{\MS}[1]{}

\newcommand{\PMJ}[1]{}
\newcommand{\pmj}[1]{#1}
\fi

\newcommand{\comment}[1]{}
\newcommand{\parag}[1]{\vspace{-3mm}\paragraph{#1}}

\newcommand{\mA}{\mathbf{A}}

\newcommand{\mL}{\mathcal{L}}

\newcommand{\mW}{\mathbf{W}} 
\newcommand{\mX}{\mathbf{X}} 
\newcommand{\mY}{\mathbf{Y}}

\newcommand{\bX}[0]{\mathbf{X}}
\newcommand{\bY}[0]{\mathbf{Y}}

\newcommand{\aswnet}[0]{\textbf{Weak-Net}}

\usepackage{multirow}

\usepackage[pagebackref,breaklinks,colorlinks]{hyperref}

\begin{document}
	\title{Weakly Supervised Volumetric Image Segmentation with Deformed Templates}
	\titlerunning{Abbreviated paper title}
	%
	\author{Udaranga Wickramasinghe\textsuperscript{1}, Patrick Jensen\textsuperscript{1,2}, Mian Shah\textsuperscript{1}, Jiancheng Yang\textsuperscript{1,3}, Pascal Fua\textsuperscript{1}}
	%
	%
	\institute{\textsuperscript{1}EPFL, Switzerland ~~ \textsuperscript{2}DTU, Denmark~~ \textsuperscript{3}Shanghai Jiao Tong University, China}
	%
	\maketitle              
	%

\begin{abstract}

There are many approaches to weakly-supervised training of networks to segment 2D images. By contrast, existing approaches to segmenting volumetric images rely on full-supervision of a subset of 2D slices of the 3D volume. We propose an approach to volume segmentation that is truly weakly-supervised in the sense that we only need to provide a sparse set of 3D points on the surface of target objects instead of detailed 2D masks. We use the 3D points to deform a 3D template so that it roughly matches the target object outlines and we introduce an architecture that exploits the supervision it provides to train a network to find accurate boundaries. We evaluate our approach on Computed Tomography (CT), Magnetic Resonance Imagery (MRI) and Electron Microscopy (EM) image datasets and show that it substantially reduces the required amount of effort. 

\end{abstract}

 %

\vspace{-5mm}
\section{Introduction} 

State-of-the-Art volumetric segmentation techniques rely on Convolutional Neural Networks (CNNs) operating on image volumes~\cite{Cicek16,Shvets18}. However, their performance depends critically on obtaining enough annotated data, which itself requires expert knowledge and is both tedious and expensive. 

Weakly-supervised image segmentation techniques can be used to mitigate this problem. They typically rely on tag annotations \cite{Huang18,Ge18a} or coarse object annotations in the form of point annotati-ons~\cite{Bearman16,Zhao20}, bounding box annotations~\cite{Hsu19}, scribbles~\cite{Zhao20} or approximate target shapes~\cite{Khoreva17}. However, these techniques have been mostly demonstrated in 2D and do not provide enough information when segmenting complex shapes, such as the liver or the hippocampus. For 3D volume segmentation, the dominant approach is to fully label a subset of 2D slices~\cite{Cicek16,Dorent20}. This is often referred to as {\it weak supervision}, even though it requires full supervision within individual slices. 


\begin{figure*}
	\centering
	\includegraphics[width=\columnwidth]{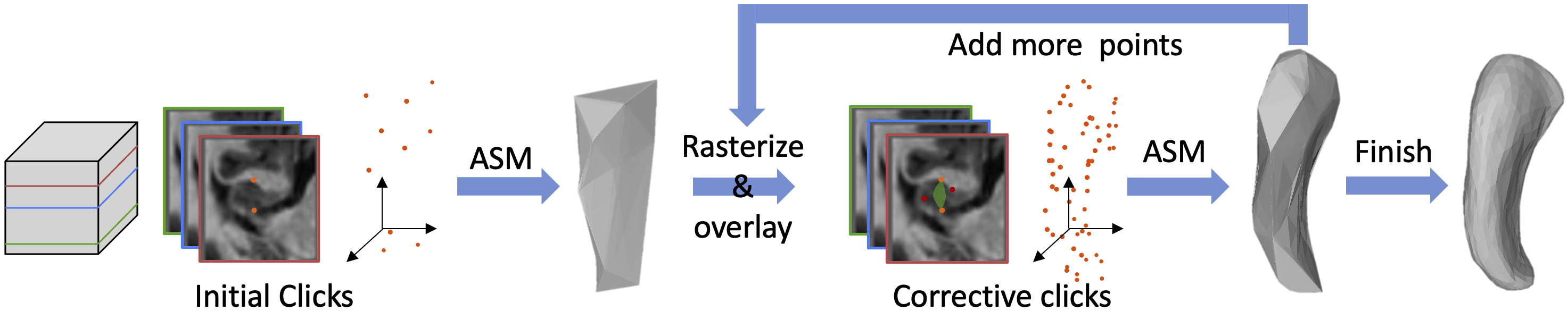}
	\vspace{-5mm}
	\caption{\small {\bf Iterative annotation strategy}. The annotator provides a few 3D points, which we fit to the template using an Active Surface Model (ASM)~\cite{Wickramasinghe21}. The result is rasterized and overlaid on the images. The annotator can then add more and deform the template again as needed.} 
	\vspace{-6mm}
	\label{fig:labelling}
\end{figure*}

\begin{figure*}
\vspace{-5mm}
\centering
\includegraphics[height=3.80cm]{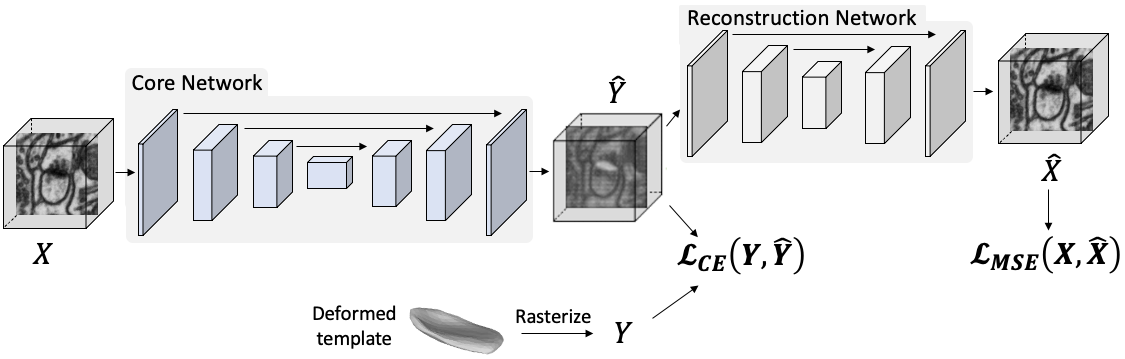}
\vspace{-3mm}
\caption{\small{\aswnet{} architecture. } A first U-Net takes the image $\mX$ as input and outputs a segmentation $\hat{\mY}$, which is in turn fed to a second U-Net that outputs a reconstructed image $\hat{\mX}$. Training is achieved by jointly minimizing  $L_{mse}$ and $L_{ce}$. This encourages $\hat{\mX}$ to resemble the original image and $\hat{\mY}$ to be similar to $\mY$, a roughly aligned version of a template.}
\vspace{-5mm}
\label{fig:arch}
\end{figure*}

By contrast, we propose a truly weakly-supervised approach that only requires a sparse set of 3D points on the surface of the target objects instead of the usual 2D masks in selected slices.  Given an appropriate user-interface, this is much faster and easier because it eliminates the need to painstakingly outline fine details, as shown in Fig.~\ref{fig:labelling}. To this end, we introduce the \aswnet{} architecture depicted by Fig.~\ref{fig:arch}.  It comprises two U-Net-like networks. The first one produces a segmentation map that matches a rough model of the target object obtained from the 3D point annotations. The second one takes the map as input and uses it to reconstruct the original image, which forces the segmentation boundaries to be accurate even though those of the template are not. 

We evaluate the performance of \aswnet{} on Computed Tomography (CT), Magnetic Resonance Imagery (MRI) and Electron Microscopy (EM) datasets. We show that it outperforms the standard approach to weak-supervision in 3D at a reduced supervision cost. More specifically, we can deliver the same accuracy as when fully annotating 2D slices for less than a third of the annotation effort. This matters because annotators typically are experts whose time is both scarce and valuable. Furthermore, it creates the basis for interactive annotating strategies that deliver the full accuracy at a lower cost than full supervision.


\vspace{-2mm}
\section{Related Work} 

We review current approaches to weak-supervision for  2D and 3D image segment- ation. We then discuss using atlases to segment biomedical image volumes. 

\subsection{Weakly-Supervised Image Segmentation}
\label{sec:weak}

\paragraph{Segmenting 2D Images.}

Tag and box annotations are among the weakest forms of annotations used to segment natural and medical images~\cite{Feng17,Huang18,Hsu19}. However, they rarely provide enough supervision for accurate results.  By contrast,  point annotations~\cite{Bearman16}, scribbles~\cite{Zhao20}, and approximate shape annotation~\cite{Khoreva17} can be used to provide useful shape information. 

The annotation process can be sped-up using dynamic programming~\cite{Mortensen95} or deformable contours~\cite{Kass88}. This makes it possible to mark only a subset of points along a contour and have the system refine it to match the target object boundary. Unfortunately, these algorithms are hard to deploy effectively in medical imagery because there are many contours besides those of interest and they can easily confuse these algorithms. This is addressed in~\cite{Peng20} by introducing {\it deep deformable contours} that only need a simple approximate contour for initialization purposes. In~\cite{Acuna18,Ling19}, the annotator is brought into the loop by giving corrective clicks when necessary. However, these deep contours require fully labelled data for their own training. 

In the result section, we show that supplying a few 3D points, as we do, is faster than using these techniques to delineate whole 2D contours in slices, as is often done.

\paragraph{Segmenting 3D Volumes.}

By far the most prevalent approach to weak-supervision for 3D segmentation, is to {\it fully} annotate subsets of slices from the training volumes~\cite{Cicek16}. Even though this reduces the segmentation effort, the annotator still has to carefully trace the object boundary in those slices. The effort can be reduced by using scribbles in individual slices~\cite{Dorent20}  or any of the semi-automated techniques described above. Unfortunately, the effectiveness of those that do not require training is limited~\cite{Mortensen95,Kass88} while the others~\cite{Acuna18,Ling19,Peng20} require full supervision and are not applicable in our scenario.

Part of the problem is that these annotation techniques operate purely in 2D without exploiting the 3D nature of the data. Because we do so, we only need a comparatively small number of point annotations for effective training. 

\vspace{-3mm}
\subsection{Template Based Approaches}
\label{sec:atlas}

Shape priors have long been used for image segmentation~\cite{Freedman05} and much recent work use shape priors in conjunction with deep nets~\cite{Wickramasinghe20,Mirikharaji18} 
For medical imaging, these priors are usually supplied in the form of sophisticated templates known as Probabilistic  Atlases (PAs) that assign to each pixel or voxel a probability of belonging to a specific class.  They are typically built by fusing multiple manually annotated images and used as auxiliary CNN inputs to provide localization priors that help the network find structures of interest. The PAs can be either very detailed, as in~\cite{Spitzer18}, to model  structures that are known in detail or very rough, as in~\cite{Wickramasinghe19}, to deal with 3D structures whose shape can vary significantly. 

PAs built by annotating points have been used in medical imaging~\cite{Januszewski18,Roth19} as a source of prior information. They are often referred to as seed layers and come in two main flavors,  Gaussian priors~\cite{Roth19}  or binary seed layers~\cite{Januszewski18}. These seed layers either indicate points inside the object~\cite{Yang18e,Januszewski18} or points on the boundary of the object~\cite{Maninis18,Wang20a}.
In the context of deep learning, PAs have been mostly used in fully-supervised approaches~\cite{Januszewski18,Roth19,Spitzer18}. An exception is the work of~\cite{Spitzer18}. However, they are only used for pre-training purposes.  Another are the one-shot and few-shot learning-based segmentation algorithms of~\cite{Dalca19}. However, they require a few fully annotated target objects as the atlases, which we do not.

\vspace{-3mm}
\subsection{Image Reconstruction}

Image reconstruction is used for semi-supervised~\cite{Liu21} and unsupervised ~\cite{Xia17} image segmentation as an auxiliary task to improve the results. In this work, we demonstrate that this idea also applies in a weakly-supervised setting to improve the segmentations produced by the core-network trained with rough annotations. In our framework, the image reconstruction network helps refine the rough initial shapes we obtain from the annotations. There are alternative approaches to boundary refinement~\cite{Acuna19} but they are designed for 2D segmentation and extending them to 3D would be non-trivial.

\section{Method}  

\label{sec:method} 

Our  approach to training a network involves an annotator providing only a sparse set of 3D points on the surface of target objects for each image volume, as opposed to carefully annotating several individual slices in each. These points are used to deform a template, such as those shown in Fig.~\ref{fig:labelling}, using active surface models~\cite{Wickramasinghe21}. This provides a rough indication of where the target object boundary is.  \aswnet{} uses it to learn weights that yield accurate object boundaries.  In short, we provide minimal human input at training time so that, at inference time, the trained network can be used without human intervention.


\vspace{-3mm}
\subsection{Network Architecture and Losses}
\label{sec:network}

\aswnet{} is depicted by Fig.~\ref{fig:arch} and comprises two separate U-Net networks~\cite{Cicek16}. The first takes as input an image volume $\mX$ of $D \times H \times W$, where $D$, $H$, $W$ stand for depth, height and width. It outputs a tensor $\hat{\mY}$ of dimension $D \times H \times W$ that stores the probabilities of each voxel belonging to the foreground.  The second takes $\hat{\mY}$ as input and yields $\hat{\mX}$, which is of the same dimension as $\mX$. Ideally, $\hat{\mY}$ should be the desired segmentation and $\hat{\mX}$ should be equal to $\mX$.  

To train \aswnet{}, we minimize a weighted sum of two losses
\begin{align}
\mL               &=  \mL_{ce} + \lambda \mL_{mse}   \; , \label{eq:total_loss} \\
\mL_{ce}      &= -\sum_{i,j,k}   \mY_{i,j,k} \log(\hat{\mY}_{i,j,k})   \; , \nonumber  \\
\mL_{mse}   &=  \sum_{i,j,k} \mW^{mse}_{i,j,k} (\mX_{i,j,k} - \hat{\mX}_{i,j,k})^2   \; , \nonumber 
\end{align}
where $\mY$ is the rasterized template fitted to the target object and $\lambda$ is a scalar that controls the influence of the second loss. $\mL_{ce}$ is a standard cross entropy loss whose minimization promotes similarity between the rasterized template $\mY$ and the segmentation  $\hat{\mY}$. $\mL_{mse}$ is a voxel-wise mean squared error in which the individual voxels are given weights $W^{mse}_{i,j,k}$ whose value is high within a distance $d$ from boundaries in $\mY$ and low elsewhere. At inference time, we only use the first U-Net, which we will refer to as our {\it core network}, and obtain the final segmentation by thresholding its output $\hat{\mY}$ using a threshold $\gamma$.

Minimizing $\mL_{ce}$ during training ensures that the segmentations will be roughly correct.  However, because the template can only be expected to provide a coarse depiction of the object, this is not enough. We therefore also minimize $\mL_{mse}$ to force the network to yield accurate boundaries. Fig.~\ref{fig:lambda} illustrates the influence of the $\gamma$ and $\lambda$ parameters on a real image. In practice, the results are insensitive  to how they are chosen over a wide range. However, minimizing  $\mL_{ce}$ alone ($\lambda$=0) produces boundaries in  $\hat{\mY}$ that are exactly those of the template while minimizing  $\mL_{mse}$ alone ($\lambda$=1000) yields boundaries that exist in the image but are not necessarily those we are looking for. Minimizing a properly weighted sum of the two yields segmentations that conform to the template while matching actual image boundaries. 


\begin{figure*}[ht]
\centering
\vspace{-4mm}
\includegraphics[width=\columnwidth]{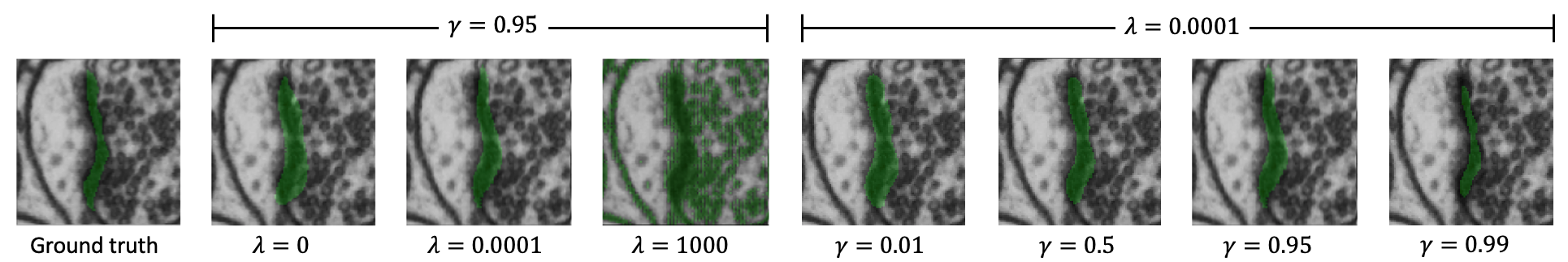}
  \vspace{-6mm}
\caption{\small {Impact of the $\lambda$ parameter in the loss of Eq.~\ref{eq:total_loss} and the thresholding parameter $\gamma$}. (Col. 2-4) When $\lambda=0$, the segmentation is very similar to the template. When $\lambda$ is large, the segmentation follows boundaries that exist in the image but are not necessarily the right ones. In between, the boundaries are correct. (Col. 5-8) We set $\lambda$ to $10^{-4}$ and vary $\gamma$. }
  \vspace{-4mm}
\label{fig:lambda}
\end{figure*}

\vspace{-7mm}
\subsection{Template Deformation}
\label{sec:deformation}


\begin{figure*}[ht!]
  \vspace{-6mm}
  \centering
  \includegraphics[height=1.95cm]{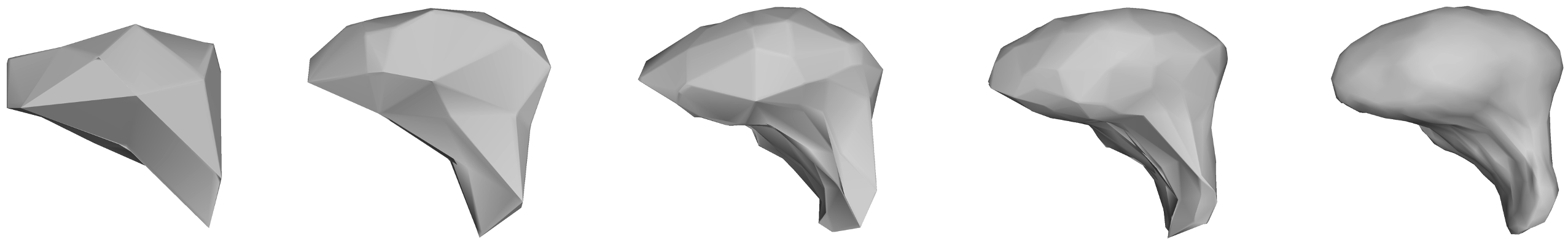}\\
  \hspace{0.3cm}  (a) \hspace{2.2cm} (b)  \hspace{2.2cm}(c) \hspace{2.0cm}(d) \hspace{2.2cm}(e) \hspace{0.0cm}
  \vspace{-2mm}
  \caption{\small {\bf Deformed templates.} (a,b,c,d,e) Deformed  template using $N=25, 50, 125, 250$, and $3661$ user-supplied 3D points when segmenting a liver. $N=3661$ corresponds to full annotations in all slices. } 
  \vspace{-5mm}
  \label{fig:template}
\end{figure*}
 
The template $\mY$ of Eq.~\ref{eq:total_loss} should approximately match the target structure. Hence, the annotator should supply points that are distributed across the object surface. These points are then used to deform the template. In practice,  For structures of genus 0, we start from a simple spherical template but more complex ones are possible. As we increase the number of points, we get increasingly refined templates, as shown in Fig.~\ref{fig:template}.

To perform this deformation interactively, we developed a GUI that relies on Active Surface Models (ASMs)~\cite{Wickramasinghe21} implemented as a MITK~\cite{Wolf2004} plugin. It lets the annotator supply a few points by clicking on 2D cross sections of the input image volume. The ASM then deforms the template in real-time and overlays it on the image data, both as 2D cross sections and 3D surface renderings. The annotator can then add more points wherever the deformed template is too far from the target organ's boundary and iterate as often as necessary. This effectively puts the human in the loop in a painless and practical way. We illustrate this in a video that can be found in the supp. material.

\vspace{-2mm}
\section{Experiments}
 \label{sec:experiment}

\vspace{-2mm}
\subsection{Datasets, Metrics and Baseline}
 \label{sec:dataset}

We use three datasets  acquired using MRI, FIB-SEM, and CT from to image the hippocampus, synaptic junctions, and the liver. We aim to produce segmentations that are above a given quality threshold using as few annotations as possible. To quantify this goal, we use two metrics, one for quality and the other for annotation effort. 

\parag{Intersection over Union.} We use the standard IoU metric \cite{Cicek16} to quantify segmentation quality.

\parag{Annotation Effort.} We use the number of points provided by the annotator as a proxy for amount of effort. When the annotator provides individual points, this is clearly proportional to the time spent. When the annotator outlines contours on 2D slices,  this becomes the sum of the contour lengths in each slice. It could be argued that this is an overestimate because the task is easier. However, in our experience it is not because precisely outlining a contour requires deliberation. 

\parag{Baseline} As discussed in Section~\ref{sec:weak}, the generally accepted way to provide weak-supervision for 3D image segmentation is to fully annotate a few 2D slices~\cite{Cicek16}. To provide a baseline, we therefore use this approach to train a 3D U-Net, as described in \cite{Cicek16}. For a fair comparison, we use the same one as in  \aswnet{}.

 
\vspace{-2mm}
\subsection{Comparative Results}

We exploit the real-time performance of active surface models to enable the annotator to provide a few points, deform the template accordingly, and then add more points where the deformed shape is not satisfactory. We first benchmark this scenario using human annotations and then provide results using simulated annotation on the object surfaces to eliminate the subjective element it contains. 

\vspace{-2mm}
\subsection{Human Annotations}

\subsubsection{Weak vs Full Supervision.}

We compare \aswnet{} and the baseline against providing full supervision and report the results in Tab.~\ref{table:results}. \aswnet{} delivers higher IoU numbers than the baseline on all three datasets. It delivers $92$ to $97\%$ of the accuracy that can be achieved with full supervision for only $7$ to $10\%$ of the annotation effort, as defined above. 


\begin{table}[ht]
	\centering
	\vspace{-4mm}
	\caption{{\bf Performance given human annotations.} Values are given as mean$\pm$std. \aswnet{} consistently outperforms the baseline in terms of both IoU and annotation effort required to achieve it. }
	\label{table:results}
	\vspace{-3mm}
	\begin{small}
		\begin{tabular}{llccc}
			\toprule
			& & Hippocampus & Liver & Syn. Junction \\
			\midrule
			Baseline & IoU (\%) & 71.2 $\pm$ 0.9 & 81.9 $\pm$ 0.7 & 68.2 $\pm$ 0.9  \\
			& Annotation Effort (\%) & 12.4\% & 6.9\% & 10.8\% \\
			\midrule
			\aswnet{}     & IoU (\%) & 74.2 $\pm$ 0.7 & 84.2 $\pm$ 1.0 & 71.5 $\pm$ 1.4 \\
			~ &  Annotation Effort (\%)  & 10.1\% & 6.8\% & 9.2\% \\
			\midrule
			\multicolumn{2}{l}{Full Annotation} & 79.3 $\pm$ 0.4 & 87.3 $\pm$ 0.3 &  73.3 $\pm$ 0.6                  \\ \hline
			\bottomrule
		\end{tabular}
	\end{small}
	\vspace{-7mm}
\end{table}

\vspace{-2mm}
\subsubsection{Annotation Time.}
\label{sec:human_annot}

In the results of Tab.~\ref{table:results}, we use the number of points supplied by the annotators to gauge the annotation effort. To complement this, we asked the annotators to manually annotate some images using other MITK tools~\cite{Wolf2004}, slice-by-slice \textit{manual contouring} of the borders, \textit{3D Region Growing}, and \textit{3D Fast Marching}. \textit{3D Region Growing} and \textit{3D Fast Marching} produce many false positive and negative regions, as shown in the supp. material. Therefore, we had to perform slice-by-slice corrections to obtain final segmentation using these two tools. In Tab.~\ref{tab:human-results}, we report the average time it took to fully annotate a single sample using each tool. As the Hippocampus volumes are small, point annotation is only $\sim$1.5x faster than annotating the full volume. For the large Liver volumes however, point annotation is $\sim$10x faster, which is significant.


\begin{table}[ht]
	\centering
	\vspace{-3mm}
	\caption{{\bf Annotation time.} Values are given as mean$\pm$std. Times are in minutes.}
	\label{tab:human-results}
	\vspace{-3mm}
	\begin{small}
\begin{tabular}{|l|c|c|}
	\hline
	~ & Hippocampus & Liver  \\ \hline
	Manual Contouring & 6.7$\pm$0.7 min. & 125$\pm$19.3 min. \\ \hline
	3D Region Growing & 5.4$\pm$1.2 min.& 75.4$\pm$30.5 min. \\ \hline
	3D Fast Marching & 5.7$\pm$1.4 min. & 86.1$\pm$25.3 min. \\ \hline
	Ours & 4.6$\pm$0.9 min. & 12.3$\pm$1.6 min.  \\  \hline
\end{tabular}
	\end{small}
	\vspace{-11mm}
\end{table}

\vspace{-2mm}
\subsubsection{Simulated Annotations}
\label{sec:simul}

To eliminate subjectivity from our experiments, we use the fact that our datasets are fully annotated to simulate the annotation process using the algorithm described in the supp. material. It is driven by two numbers, $N$ the number of points per sample and $P$ the number of samples we annotate. To keep the number of experiments within a manageable range, we vary both $N$ and $P$ when experimenting on the Hippocampus dataset and only $N$ for the other two. To provide a baseline, we randomly pick a number of slices from three image planes to be annotated and use the ground-truth annotations for these slices. When selecting them, we check that they contain the target object. When evaluating the baseline, we vary the number of slices we use and of samples we annotate.


\begin{figure*}[t!]
\centering
\vspace{-3mm}
\includegraphics[width=0.995\linewidth]{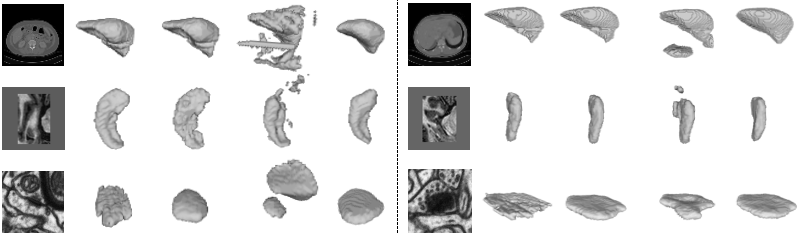}
\raggedright\hspace*{0.2cm}  (a) \hspace{0.6cm} (b) \hspace{0.6cm} (c) \hspace{0.6cm} (d) \hspace{0.6cm} (e) \hspace{0.6cm}  (f) \hspace{0.6cm} (g) \hspace{0.6cm} (h) \hspace{0.6cm} (i) \hspace{0.6cm} (j)
\vspace{-3mm}
\caption{\textbf{Segmentation results.} (a+f) A slice from the input volume from top to bottom (b+g) Ground truth (c+h) U-Net with full-supervision (d+i) Corresponding Baseline (e) \aswnet{} trained with 25 points per sample. (f) \aswnet{} trained with 125 points per sample.}
\vspace{-6mm}
\label{fig:results}
\end{figure*}

\begin{figure}[h]
\vspace{-4mm}
	\centering
	\begin{tabular}{cc}
	 \includegraphics[width=0.40\linewidth]{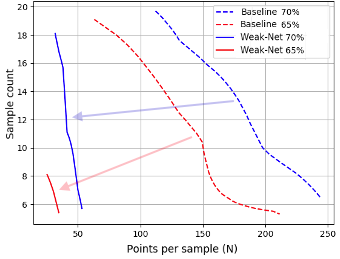} 
	 \begin{tabular}{c}
	  \includegraphics[width=0.35\linewidth]{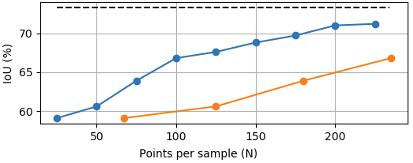} \\
	  \includegraphics[width=0.35\linewidth]{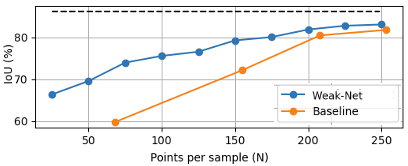} \\
	  \vspace{30mm}
	 \end{tabular}
	 \vspace{-37mm}
	 \end{tabular}
\caption{\textbf{Performance given synthetic annotations} (left) Annotation effort required to achieve $65\%$ and $70\%$ IoU on the hippocampus. The blue and red arrows indicate the effort reduction our approach delivers. (right) IoU as function of the number of points used (N) for the synaptic junction and liver dataset sets. The dashed line denotes fully supervised results.}
\vspace{-8mm}
\label{fig:plot3}
\end{figure}

We present qualitative results of the experiments in Fig.~\ref{fig:results}. On the left of Fig.~\ref{fig:plot3}, we plot the total annotation effort, that is, the product of the number of points per sample and the number of samples, required to attain a target IoU---either $65\%$ or $70\%$---in the Hippocampus dataset. As there are many ways to achieve a given IoU by increasing one number while decreasing the other, we draw {\it iso-IoU} curves. The baseline ones are dashed and ours are full and clearly to the left of the dashed ones. In other words, we need significantly less effort to achieve a similar result. On the right of Fig.~\ref{fig:plot3}, we report our results on the other two datasets that contain fewer samples. Hence we used all training samples and plot the IoU as a function of the number of points per sample. For the same number of points, our approach consistently outperforms the baseline. When using it, we cannot annotate less than one slice and, hence, reduce the annotation effort below a certain level, which is why the blue curves extend further to the left than the yellow ones. 


\section{Conclusion}
 We have presented a weakly supervised approach to segmenting 3D image volumes that outperforms more traditional approaches that rely on fully annotating individual 2D slices. 
 
It relies on deforming simple spherical templates that incorporate no shape prior. In future work, to further reduce the annotation burden, we will develop more sophisticated templates that are parameterized in terms of low-dimensional latent vectors and can therefore be deformed by specifying even fewer 3D points than we do now.

\section{Appendix}




\begin{figure}[h]
\centering
\includegraphics[height=5.5cm]{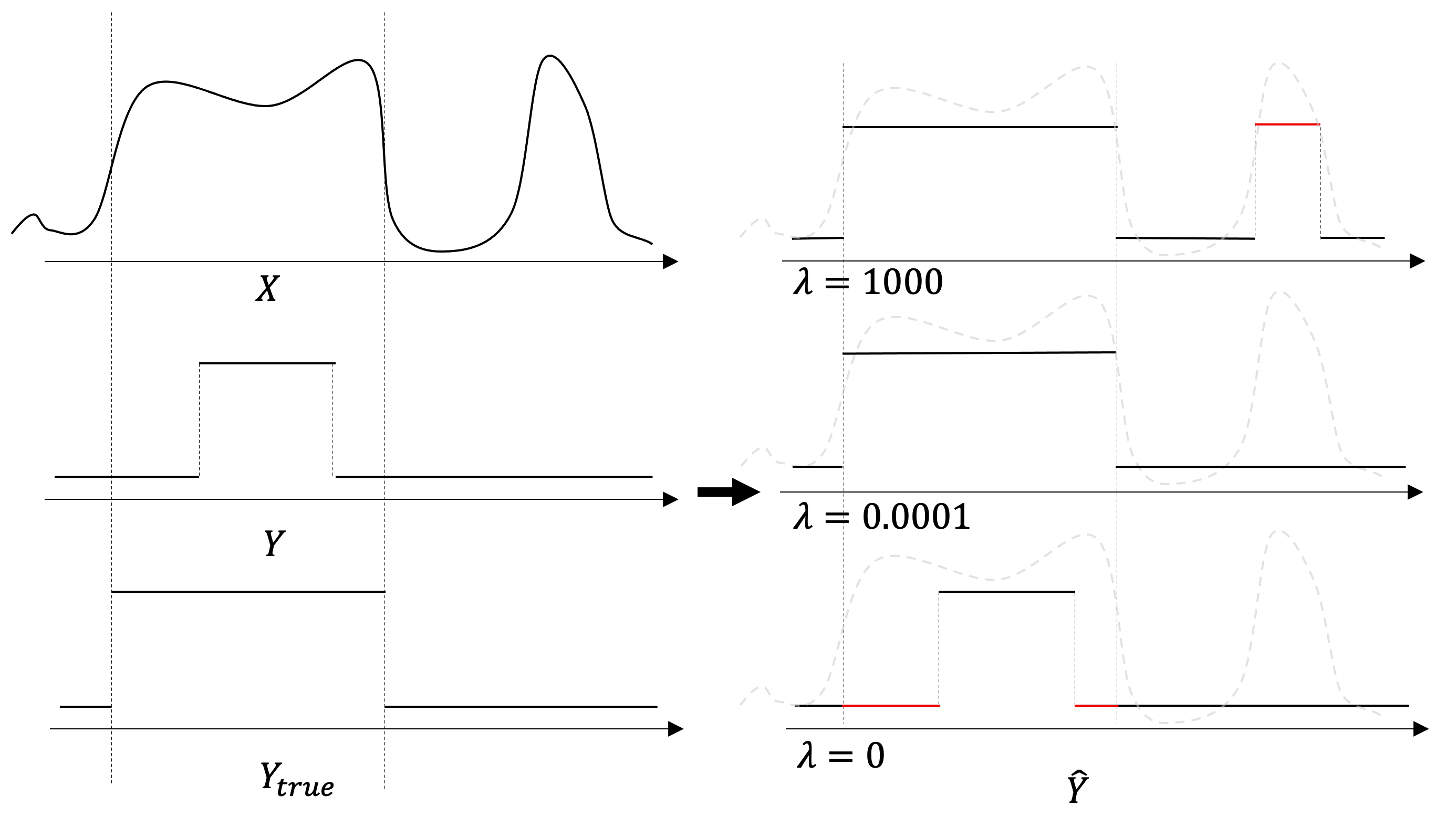}\\
(a) \hspace{4.5cm} (b)
\vspace{-2mm}
\caption{\small {\bf Using a secondary task to improve segmentations}. (a) $\bX$ is the input and $\bY_{true}$ is the ground-truth. $\bY$ is the atlas that is roughly correct. (b) When $\lambda$ is too large, $\mL_{mse}$ dominates and the resulting segmentation features a spurious region shown in red that corresponds to image boundaries that are {\it not} those of the target structure. When $\lambda$ is zero, only $\mL_{ce}$ is minimized and minimizing produces boundaries that are also those of the atlas, which are not at the right place as the denoted by the red lines. For appropriate values of $\lambda$, the boundaries are correct and the spurious region is eliminated.}\label{fig:sketch}
\end{figure}


\begin{figure}[h]
\centering 
	\centering
	\includegraphics[height=2.3cm]{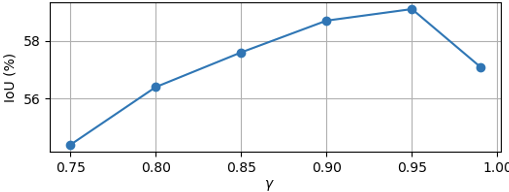} 
	\includegraphics[height=2.3cm]{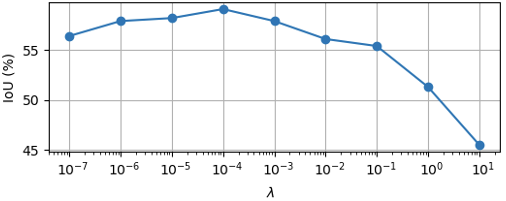} 
	  
\caption{ \textbf{Influence of $\lambda$ and $\gamma$ parameters} . IoU (\%) as a function of $\lambda$ in (a), $\gamma$ in (b) on the Synaptic Junction dataset with $N=25$.   }
\label{fig:lambda}
\end{figure}

 
 \subsection{Sampling Algorithm}
\begin{figure}[t!]
\centering\
	\centering
	\includegraphics[height=3.0cm]{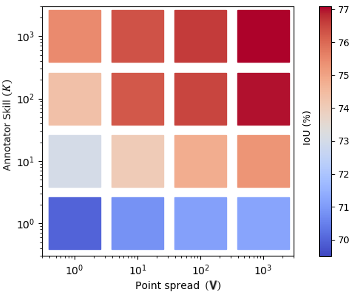} 
	~
	\includegraphics[height=3.0cm]{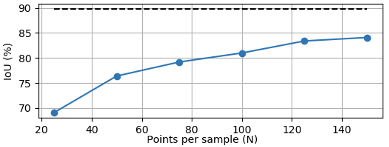} 
	
\vspace{-2mm}
\caption{\textbf{Annotation simulation on the Liver dataset} (a) Reconstruction accuracy (IoU) variation for different values of $K$ and $V$.  (b) Reconstruction accuracy (IoU) variation as a function of the number of annotated points. The black-dashed line indicates the accuracy that would be achieve by annotating all points in each sample, that is 3661 points per sample on average for the liver dataset. }
\label{fig:anno_simulation}
\end{figure}

Simulated annotations are generated using the \textbf{Improved selection} strategy which is based on \textbf{Random selection} strategy.
\begin{enumerate}

	\item {\bf Random selection.} $N$ points are randomly selected across the 3D surface. Ideally, they should be uniformly distributed across the surface but there is no guarantee of that. To emulate the behavior of a conscientious annotator trying to achieve this, we repeat the operation $V$ times and select the set of $N$ points that exhibits the largest intra-point variance. As we increase $V$, so does the probability that the selected points will indeed be uniformly distributed. The influence of variables $N$ and $V$ is demonstrated in Fig~\ref{fig:anno_simulation}.

	\item {\bf Improved selection.} We simulate the fact that a skilled annotator will provide the most informative points possible by performing $K$ random annotations as described above, using each one to deform the template, and selecting the one that yields the highest IoU with the ground truth. As $K$ increases, so does the probability that the deformed template will match the ground truth well.

\end{enumerate}
%
 

\begin{figure}[h]
	
	\centering
	\includegraphics[width=0.25\linewidth]{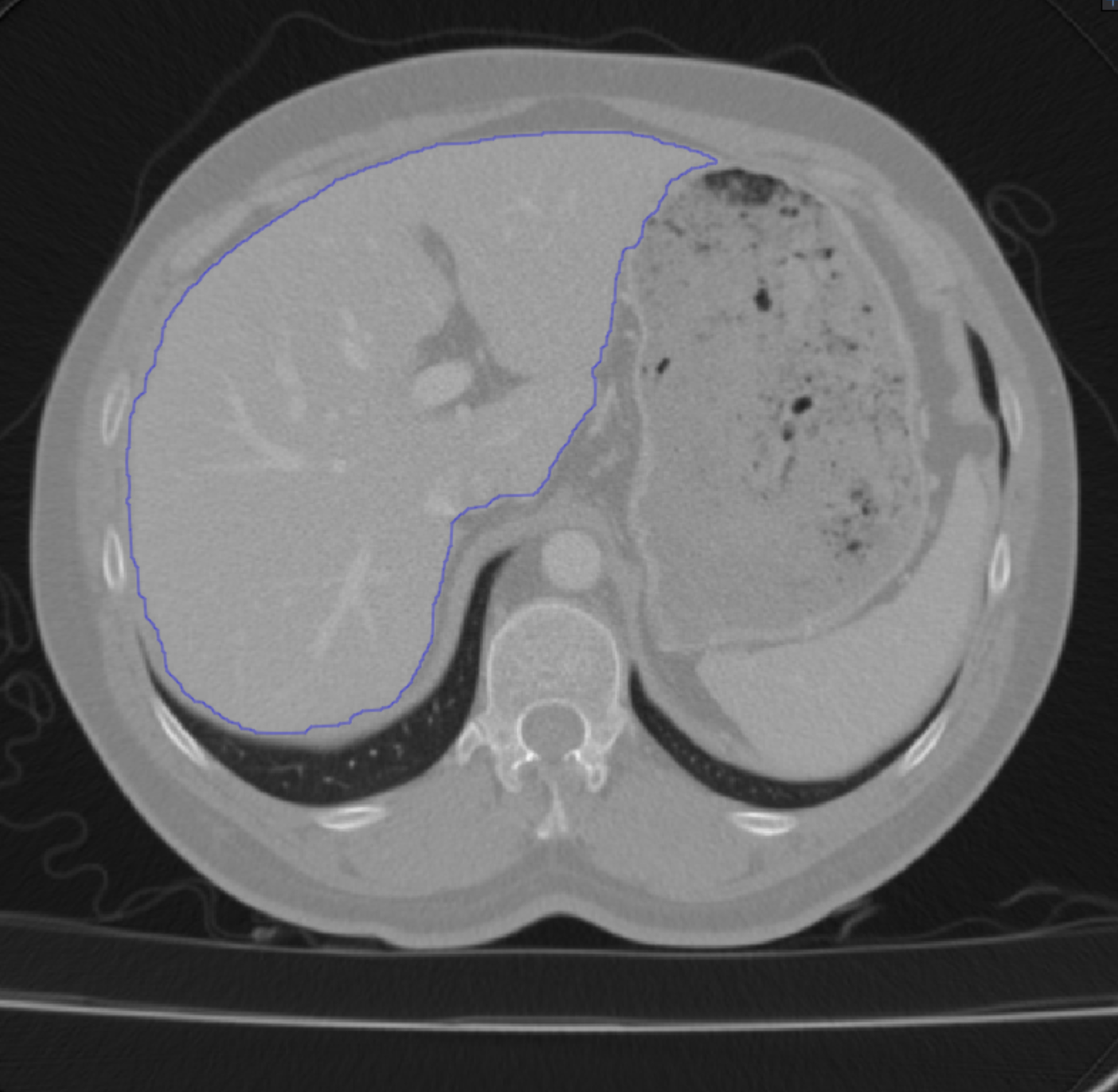} \hspace{2mm} 
	\includegraphics[width=0.25\linewidth]{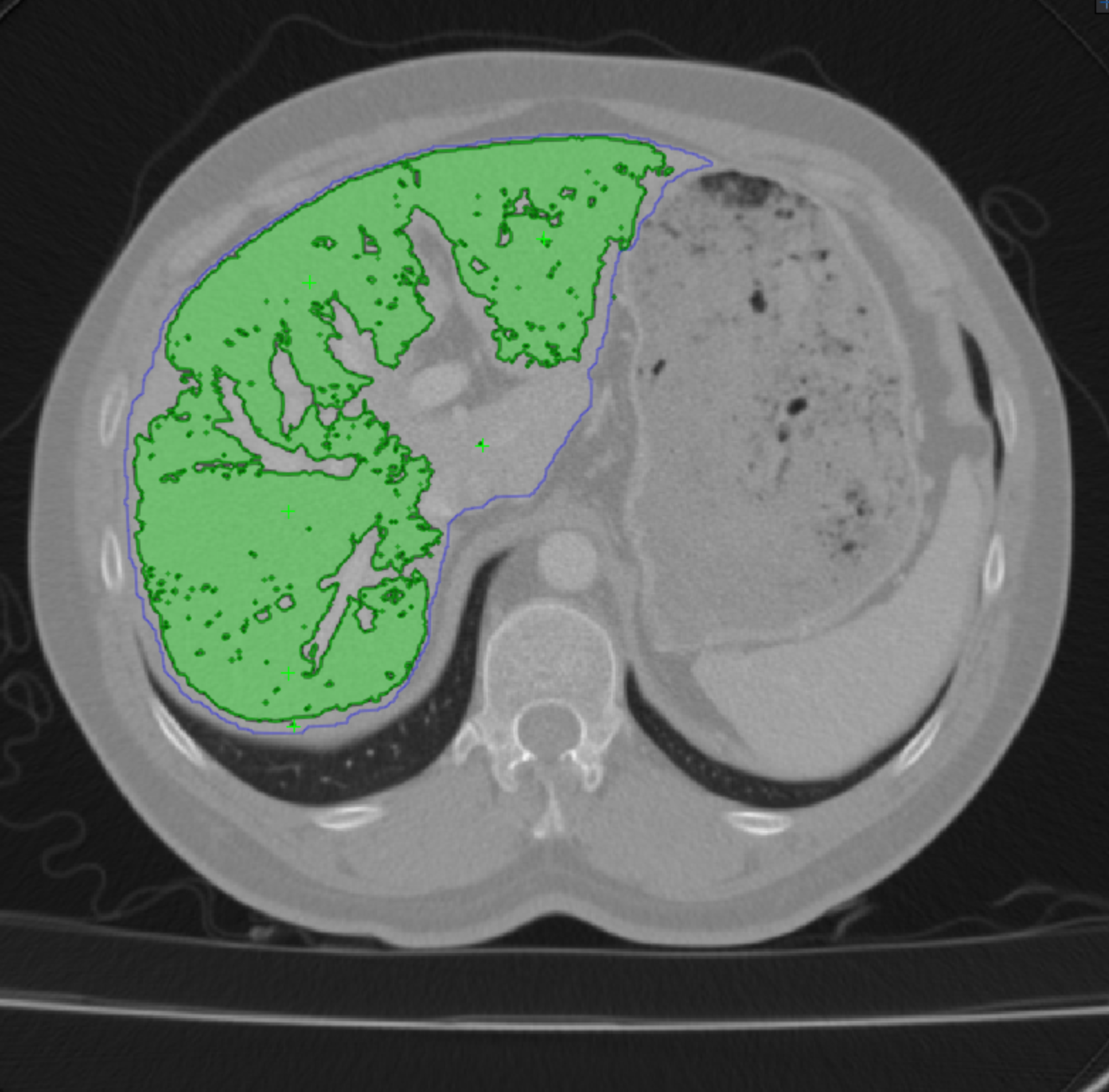}
	\hspace{2mm} 
	\includegraphics[width=0.25\linewidth]{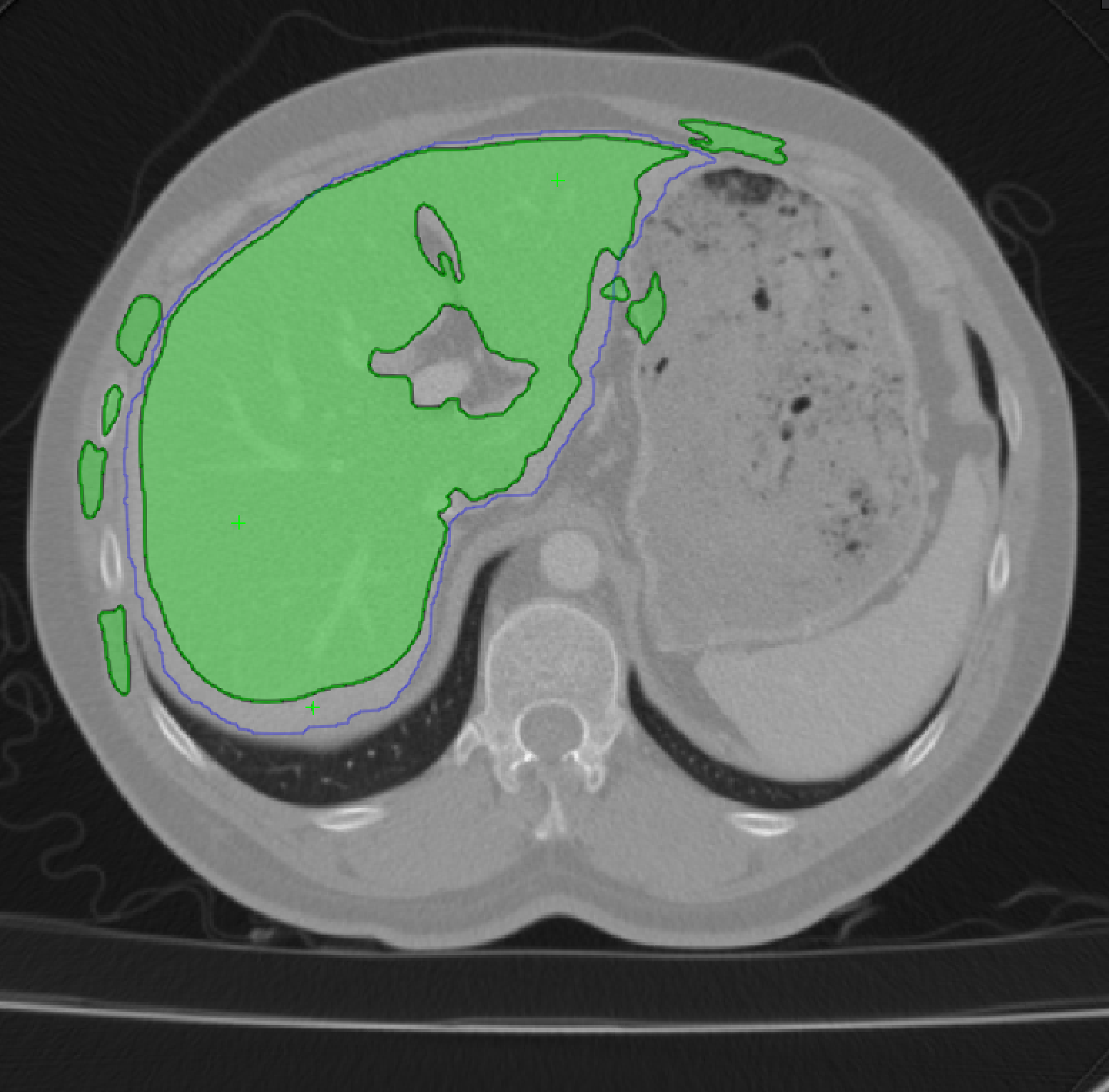}
	
	\vspace{-1mm} 
\caption{\textbf{Use of Fast Marching algorithm from MITK to annotate a liver} (a) Image with ground truth contour (in blue)  (b) Fast Marching algorithm with small $\sigma$  (c) Fast Marching algorithm with large $\sigma$. In both instances, after the initial segmentation, we have to fix false positive and false negative regions.   }
	\label{fig:fast_marching}
	
\end{figure}

\comment{

To perform this deformation interactively and in real-time, we developed the GUI interface depicted by Fig.~\ref{fig:mitk-gui}, which relies on a recent approach to Active Surface Models (ASMs)~\cite{Wickramasinghe21} that makes full-use of GPUs and is therefore fast. It is implemented as a\pmj{n} MITK~\cite{Wolf2004} plugin. It lets the annotator supply a few points by clicking on 2D cross sections of the input image volume. The ASM then deforms the template, which is overlaid on the image data, both as 2D cross sections and a 3D surface rendering. The annotator can then add more points wherever the deformed template is too far from the target organ's boundary and iterate as often as necessary. This effectively puts the human in the loop in a painless and practical way. We illustrate this in a video that can be found in the supplementary material. 
}

  \comment{
 \subsection{More Qualitative Results}

We show more qualitative results in Fig.~\ref{fig:results_more} which further demonstrate the effectiveness of our approach. Here, the results are shown from the experiments that use the minimum amount of annotation for baseline and \aswnet{}. For the baseline, it is 1 slice per sample and for \aswnet{} it is 25 points per sample. One slice on average contain 63 points in the hippocampus dataset, 67 in the synaptic junction dataset and 68 in the liver dataset.  }

 \comment{
 	
 	\subsection{Training and  Hyper-parameter tuning}
 	Experiments were conducted on 3 NVIDIA Tesla V100 GPUs. Each model was optimized using Adam \cite{Kingma14a} optimizer with a learning rate of $10^{-4}$ for 150000 iterations.  Due to the large number of experiments required to generate the plots in Fig.~\ref{fig:plot1},~\ref{fig:plot2},~\ref{fig:lambda}, only one instance of each experiment was conducted. For Table~\ref{table:ablation_results}, each experiment was repeated 3 times.
 	
 	Hyper-parameter ($\lambda$ and $\gamma$) tuning were done using the validation set of the Hippocampus dataset. As mentioned in section~\ref{sec:dataset}, same hyper-parameters were used for all three dataset because the hyper-parameters from the hippocampus dataset were sufficient to demonstrate the effectiveness of the proposed method in the other two datasets as well.

	\subsection{Template Deformation}
	\label{sec:deformation}
	

\begin{figure*}[ht!]
  \vspace{-6mm}
  \centering
  \includegraphics[height=1.95cm]{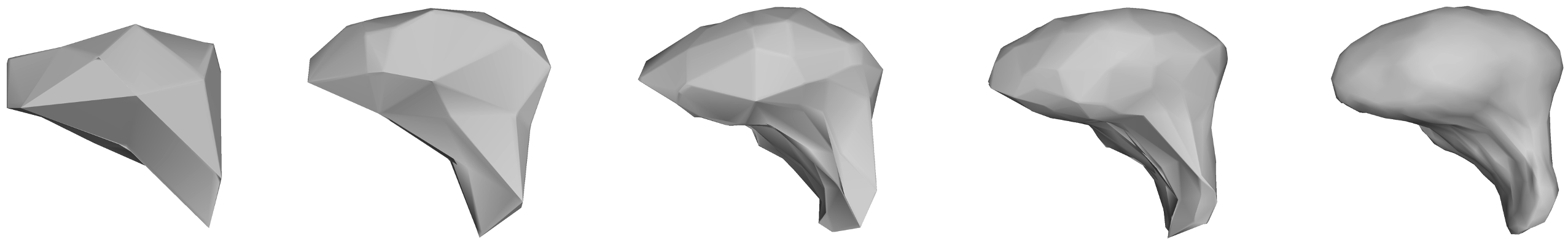}\\
  \hspace{0.3cm}  (a) \hspace{2.2cm} (b)  \hspace{2.2cm}(c) \hspace{2.0cm}(d) \hspace{2.2cm}(e) \hspace{0.0cm}
  \vspace{-2mm}
  \caption{\small {\bf Deformed templates.} (a,b,c,d,e) Deformed  template using $N=25, 50, 125, 250$, and $3661$ user-supplied 3D points when segmenting a liver. $N=3661$ corresponds to full annotations in all slices. } 
  \vspace{-5mm}
  \label{fig:template}
\end{figure*}
	Recall from Section~\ref{sec:network}, that the template $\mY$ of Eq.~\ref{eq:total_loss} should approximately match the target structure. Hence, the annotator should supply points that are distributed across the object surface. These points can then be used to deform the template so that its boundaries roughly correspond to those of the target structure. In practice, \pmj{structures of genus 0 are the most common. For these, we start from a simple spherical template but more complex ones are possible, \textit{eg} those discussed in Section.}
	We do this because creating complex atlases would require fully annotated data and it would go against the objectives of this work. As we increase the number of points, we get increasingly refined templates, as shown in Fig.~\ref{fig:template}.
}

	%
	%
	%
	
	{\small
	\bibliographystyle{splncs04}
	\bibliography{string,vision,biomed,learning,optim}
}
	%
	

\end{document}